# A Novel Paradigm for Calculating Ramsey Number via Artificial Bee Colony Algorithm[I]

Wei-Hao Mao[a], Fei Gao[a,*], Yi-Jin Dong[a], Wen-Ming Li[a]

[a]*Department of Mathematics, School of Science, Wuhan University of Technology, Luoshi Road 122, Wuhan, Hubei, 430070, People's Republic of China*

**Abstract**

The Ramsey number is of vital importance in Ramsey's theorem. This paper proposed a novel methodology for constructing Ramsey graphs about R(3,10), which uses Artificial Bee Colony optimization(ABC) to raise the lower bound of Ramsey number R(3,10). The $r(3,10)$-graph contains two limitations, that is, neither complete graphs of order 3 nor independent sets of order 10. To resolve these limitations, a special mathematical model is put in the paradigm to convert the problems into discrete optimization whose smaller minimizers are correspondent to bigger lower bound as approximation of inf R(3,10). To demonstrate the potential of the proposed method, simulations are done to to minimize the amount of these two types of graphs. For the first time, four $r$(3,9,39) graphs with best approximation for inf R(3,10) are reported in simulations to support the current lower bound for R(3,10). The experiments' results show that the proposed paradigm for Ramsey number's calculation driven by ABC is a successful method with the advantages of high precision and robustness.

[I]The work is partially supported by Supported by the State Key Program of National Natural Science of China(Grant No.91324201), the National Natural Science Foundation of China (Grant No.81271513) of China, the Fundamental Research Funds for the Central Universities of China, the self–determined and innovative research funds of WUT No. 2014-Ia-012, 2013-Ia-040, national college students' innovative entrepreneurial training program(WUT No. 20151049714003) , the Natural Science Foundation No.2014CFB865 of Hubei Province of China

*Corresponding author
*Email addresses:* mwhaea123456@163.com (Weihao Mao), gaofei@whut.edu.cn (Fei Gao),
227173@whut.edu.cn (Yijin Dong), 1206047725@qq.com (Wenming Li)

---

**1. Introduction**

The Ramsey number R($m,n$) is defined to answer the problems to invite the least number of guests in which at least m persons know each other or at least n persons don't know each other[1]. To state the definition in graph language, that should be the minimum number of vertices $v$ = R($m,n$) so that all undirected simple graphs of order $v$ contain a clique of order $m$ or an independent set of order $n$[1]. And the universal Ramsey number R($k_1,k_2,\cdots,k_r$) is the smallest integer n such that in any $r$-coloring of the edges of the complete graph $K_n$, there is a monochromatic copy of $K_{k_c}$ for some $1 \leq c \leq r$. Up to now, a lot of efforts have been made to study the Ramsey number, and there has been some valid and useful results. It is evident that R($m,n$) = R($n,m$) and R($m,2$) = $m$[2]. And Erdos[3, 4, 5] proved that for diagonal Ramsey number R($k,k$),

$$\frac{k2^{k/2}}{e\sqrt{2}} < \boldsymbol{R}(k,k)$$

In 1995, Kim obtained a breakthrough result by establishing the exact asymptotic of R(3,$k$) using probabilistic arguments[6]. To get the new upper bounds on R(3,$k$), Jan Goedgebeur[2] calculated the computation of the exact values of $e(3,k,n)$, which is devoted to be the minimum number of edges in any triangle-free graph on $n$ vertices without independent sets of order $k$, and got good results. Geoffrey Exoo has got new lower bounds for 28 classical two and three color Ramsey numbers by using two color constructions[7, 8]. Hiroshi Fujita has improved the lower bound of Ramsey R(4,8) from 56 to 58 using a SAT solver based on MiniSat and customized for solving Ramsey problems in 2012[9]. Some latest results about Ramsey number can be shown in Table 1 below [10, 11, 12, 13, 14, 15, 16, 17, 18].

Table 1: Some latest results about Ramsey numbers

| $m$ | $n$ | R($m,n$) | Reference |
|---|---|---|---|
| 3 | 3 | 6 | Greenwood and Gleason 1955[19] |
| 3 | 4 | 9 | Greenwood and Gleason 1955[19] |
| 3 | 5 | 14 | Greenwood and Gleason 1955[19] |
| 3 | 6 | 18 | Graver and Yackel 1968[20] |
| 3 | 7 | 23 | Kalbfleisch 1966[21] |
| 3 | 8 | 28 | McKay and Min 1992[22] |
| 3 | 9 | 36 | Grinstead and Roberts 1982[23] |
| 3 | 10 | [40,42] | Exoo 1989 and Goedgebeur and Radziszowski 2012[24, 25] |
| 3 | 11 | [46,50] | Goedgebeur and Radziszowski 2012[2] |
| 4 | 4 | 18 | Greenwood and Gleason 1955[19] |
| 4 | 5 | 25 | McKay and Radziszowski 1992[26] |
| 4 | 6 | [36,41] | Exoo 2012[25] |
| 4 | 8 | [59,84] | Exoo 2015[7] |
| 5 | 5 | [43,49] | Mckay and Radziszowski 1995[27] |
| 5 | 10 | [149,442] | Exoo 2015[7] |

It is obviously that $r(3,10)$–graph contains two limitations, that is, neither complete graphs of order 3 nor independent sets of order 10. To calculate the Ramsey number R(3,10), Jan Goedgebeur and Stanislaw P.Radziszowski got new Computational Upper Bounds for Ramsey Numbers R(3,$k$) in 2013[2].They have proved that $40 \leq R(3,10) \leq 42$ and anticipated that any further improvement to either of the bounds will be very difficult[2]. Despite many attempts by Exoo, them, and others, nor $r(3,10,40)$–graphs were constructed[28, 29, 30, 31, 32].

To resolve these limitations, a special mathematical model is put in the paradigm to convert the problems into discrete optimization whose smaller minimizers are correspondent to bigger lower bound of R(3,10)[33, 34, 35, 36]. And Artificial Bee Colony optimization(ABC)[37, 38, 39, 40, 41, 42, 43, 44, 45, 46, 47, 48] is to raise the lower bound of Ramsey number R(3,10).

The rest is organized as follows. Section 1 give a simple review on Ramsey number. In Section 2 some current results on Ramsey number were given. In Section 3, a novel united mathematical model for computing Ramsey number via ABC are proposed in a new methodology which converts the problems into discrete optimization whose smaller minimizers are correspondent to bigger lower bound of R(3,10). In section 4, simulations are done, And or the first time, four $r(3,9,39)$ graphs are reported in simulations to support the current lower bound for R(3,10). The experiments' results show that the proposed paradigm for Ramsey number's calculation driven by ABC is a successful method with the advantages of high precision and robustness. Conclusions are summarized briefly in Section 5.

**2. Some Results**

*2.1. The Range of degree*

**Theorem 1.** *In a Ramsey graph of $r(p,q)$ with n vertices,the degree of every vertex is in the range of $[n - r(p,q - 1), r(p - 1,q) - 1]$[27].*

Proof. For any vertex $u$, we suppose the degree of $u$ is $d$. Then there are $d$ vertices adjacent to $u$.

- Thus $d < r(p - 1,q)$. If not $u$, then this $d$ vertex will contain a $p - 1$ ₅₅ complete graph. With the addition of $u$, there will be a $p$ complete graph.

- Thus $n - 1 - d < r(p,q - 1)$. If not $u$, and $n - 1 - d$ vertices which are not adjacent to $u$ will form a $q$ dependent set.

So $n - 1 - r(p,q - 1) < d < r(p - 1,q)$. Since all of degrees are integer so $d$ is in the range of $[n - r(p,q - 1), r(p - 1,q) - 1]$.

From this conclusion we know some useful results as Table 2:
This is a very important conclusion in the further research.

Table 2: Degree of some Ramsey Graphs

| Ramsey Graph | Degree |
|---|---|
| $r(3,10,40)$ | [4,9] |
| $r(5,5,43)$ | [18,24] |
| $r(4,6,36)$ | [11,24] |

*2.2. The Uniqueness of $r(3,9,35)$*

The $r(3,9,35)$-graph was found by Kalbfleisch[21] in 1966, and in 2012 Goedgebeur and Radziszowski[2] proved the uniqueness of this graph. This graph is a 8-regular graph which is used to construct the $r(3,10,39)$ graphs.

*2.3. The Construction of $r(3,10,40)$ from $r(3,9,35)$*

If we want to add 5 vertices to $r(3,9,35)$ to construct a graph of $r(3,10,40)$.

• At first, these 5 vertices have an induced subgraph. This subgraph cannot contain a complete graph of 3. So the edges among this 5 graph are easy to identify. After this choosing process, only left 14 graphs of vertices 5.

• Since the $r(3,9,35)$ is a 8-regular graph, and the degree of vertices in $r(3,10,40)$ is in the range of [4,9], so this 5 new vertices can have edges adjacent to different vertices in $r(3,9,35)$. If not, there will be a vertex $u$ in $r(3,9,35)$ with 10 degree in this new graph.

**3. Mathematical Model for Calculating Ramsey Number via Artificial Bee Colony Algorithm**

This paper proposed a novel methodology for constructing Ramsey graphs about R(3,10), which uses Artificial Bee Colony optimization(ABC) to approximate Ramsey number's inf R(3,10). The biggest difficulties here we have resolve to be resolve is that The *r*(3,10)–graph contains two limitations, that is, neither complete graphs of order 3 nor independent sets of order 10.

And to resolve these limitations, a special mathematical model is put in the paradigm to convert the problems into discrete optimization whose smaller minimizers are correspondent to bigger lower bound as approximation of R(3,10)'s infimum as following flowchart1 shows. And the whole mathematical model will be interpreted in this Section in detail.

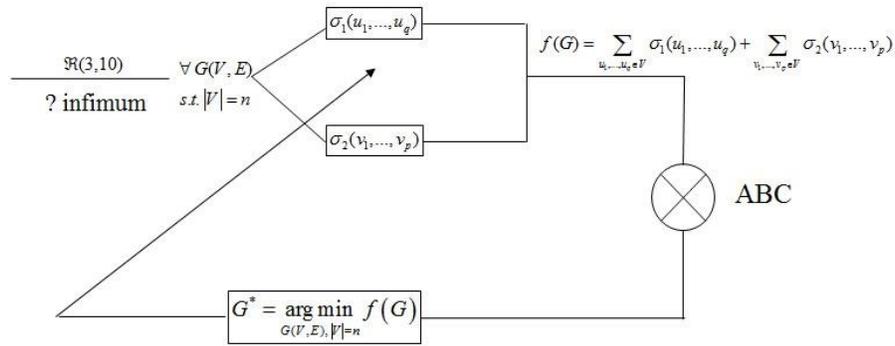

Figure 1: A novel methodology for constructing Ramsey graphs via ABC

*3.1. The objective function*

Let

$$\sigma_1(u_1, ..., u_q) = \begin{cases} 1, if\ subgraph\ induced\ by\ (u_1, ..., u_q) form\ a\ q-independent\ sets \\ 0, otherwise \end{cases}$$

$$\sigma_2(v_1, ..., v_p) = \begin{cases} 1, if\ subgraph\ induced\ by (v_1, ..., v_p)\ form\ a\ p-complete\ graph \\ 0, otherwise \end{cases}$$

Then in *r(p,q,n)* calculation, we define the fitness function of graph *G(V,E)*,|*V*| = *n* for ABC as following.

$$f(G) = \sum_{u_1,...,u_q \in V}^{q} \sigma_1(u_1,...,u_q) + \sum_{v_1,...,v_p \in V}^{p} \sigma_2(v_1,...,v_p)$$

And the objective of the ABC is *F = minf(G),(forallG(V,E),|V| = n)*.

If Graph $G$ is a graph with n vertices and with no $p$-complete graphs or q-independent sets, it means $f(G) = 0$. If Graph $G$ is a graph with $n$ vertices and with no $p$-complete graphs or $q$-independent sets, it means $f(G) = 0$.

For example, if we want to calculate Ramsey number $r(3,3,5)$, then graph ($G_1$) in Figure 1.

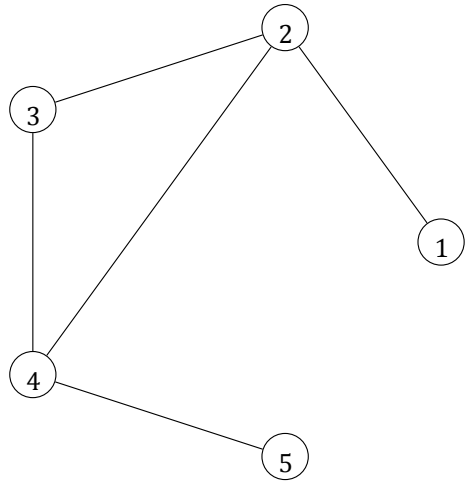

Figure2: Example Graph:G1

It's easy to see that vertice(2,3,4) form a 3-complete graph, and vertices(1,3,5) form a 3-independent sets. So $f(G_1) = 2$.

And for graph $G_2$ in Figure 2:

So it's easy to see $f(G_2) = 0$, $G_2$ is what we want in our search.

So the process to find $r(p,q,n)$ is converted into a function optimization problem. we will optimize the function via Artificial Bee Colony which is shown by the figure below.

*3.2. Adjacent Graph*

We also define that if two graphs $G_1(V_1,E_1), G_2(V_2,E_2)$ is adjacent, it means $|E_1 - E_2| = 1$, and there is only one different edge between this two graphs.

We use the adjacent matrix to represent a graph $G$. It means every bee is coded by a vector ($n \times n$ dimension 0-1), so two adjacent graphs are represented by two vectors and distinguished by only 2 different dimensions.

Still the former example $G_1$

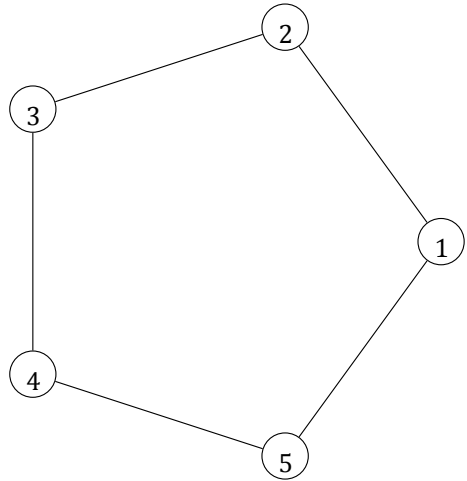

Figure3: ExampleGraph:G2

the adjacent matrix is

01000

10110

01010

01101

00010

And one of its adjacent graph $G_3$ is: its

adjacent matrix is:

01000

10100

01010

00101

00010

So there is only one different edge between $G_1 and G_3$

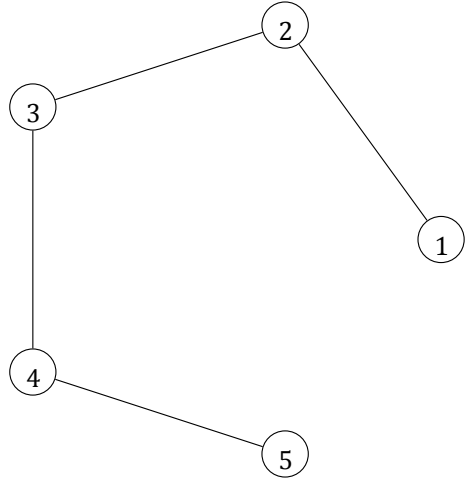

Figure4:ExampleGraph:G3

*3.3. Constructing r(3,10,40)*

*3.3.1. The Construction of A New Graph*

At first we have a 35-*r*(3,9,35) graph. As pointed out above, there are only 14 situations among these 5 new vertices.

We suppose the 35 vertices in r(3,9,35) as $v_1, v_2...v_{35}$ and its edge set is $E_r$. And suppose the new 5 vertices as $u_1, u_2, u_3, u_4, u_5$. Choose one of the 14 situations pointed out above. Then the construction of *r*(3,10,40) is among $u_1, u_2, u_3, u_4, u_5$. We suppose they are $E_1, E_2,...E_{14}$, as sets of edges. And we choose $E_p$, suppose $t_i$ as the edges adjacent to $u_i$ in $E_P$

let

$$deg(u_i) = rand\{4,5,6,7,8,9\}$$

and randomly born a permutation of 1..35 as $i_1, i_2,...i_{35}$, let

$$E = \{(u_k, i_q) | \sum_{n=1}^{k} \deg(u_n) - t_n \geq q \geq \sum_{n=1}^{k-1} \deg(u_n) - t_n\}$$

$V = \{v_1, v_2...v_{35}\} \cup \{u_1, u_2, u_3, u_4, u_5\}$

Thus we constructed a new graph $G = (V,E)$ for example, if we born a permutation as:

1,2,3,5,4,6,7,9,8,10,11,13,14,15,12,16,17,18,20,19....

and the randomly given degree is (5,8,7,6,6), so vertex (1,2,3,5,4) will be linked an edge to the vertex 36, and vertices (6,7,9,8,10,11,13,14) will be linked an edge to the vertex 37, and so on.

We choose $n$ as the amount of bees in colony. we just need to construct n new graphs.

This means that we can firstly find out $k$–independent sets in $r(3,9,35)$ and $(10 - k)$–independent sets in the five new vertices and then combine them together to see whether the subgraph induced by them can form a 10–independent sets. In this way we can save many time in fitness function calculations.

This is the independent sets contain in r(3,9,35)

Table 3: The amount of independent sets

| $r$-independent sets | The number |
| --- | --- |
| 5-independent sets | 20265 |
| 6-independent sets | 22995 |
| 7-independent sets | 13760 |
| 8-independent sets | 3360 |

We firstly save all of these $k$-independent sets, and when we calculate the fitness function, we only need to list the 10-$k$ independent sets and delete these which doesn't construct a 10-independent set. It is an efficient way to calculate every graph's fitness function. We only need to visit all the vertices adjacent to the added vertices and calculate the amount of independent sets they contain.

*3.4. Artificial Bee Colony Optimization*

In this paper we use Basic Artificial Bee Colony[40, 41, 42, 43, 44, 45, 46, 47, 48] to solve the Mathematical Model, which includes several basic processes:

Step 1. **Initialization phase.**

Calculate all the all the matters we need according to 3.1. In this process, a bee colony is produced and all of the bees are sent out to search for food source.

We first construct the first generation of graphs for every bee we set according to 2.3. After that, calculate all the fitness function of these graphs, and only left half of the bees as employed bees, the others give up their graphs and become onlooker bees. Set all the staynum(the total times it stay at the same graph) of employed bees to be 1.

Step 2. **Employed bees phase.**

All the employed bees and their followed bees are sent out to search for new graphs in their adjacent graphs according to 3.2. If the new graph is better than the old one, then change their graphs and their staynum are sent to be 1; if not they will stay at the same graph, and their staynum will +1;

If any employed bees and its followed bee stay in the same graph for many times, and staynum maxlimit then it and its followed bee will give up the present graphs and become scout bee and onlooker bee.

Step 3. **Onlooker bees phase.**

Sort the employed bee according to their fitness value [47, 48]. If there are $n$ bees, the minimum bee will be assigned a weight $w_k$ of $n$, next is $n-1$ and so on. And every onlooker bee has possibility to choose one of the employed bees which do not have a followed bee. The selection operator [49, 50, 51, 52] for a follow bee $i$ choose employed bee $j$ is:

$$P_i = \alpha \frac{w_j}{\sum_k w_k}, bee \quad k \quad do \quad not \quad have \quad a \quad folowed \quad bee$$

Step 4. **Scout bees phase.**

All the scout bee is sent out to construct a new graph and their staynum are sent to be 1.

 **4. Simulation**

After running program, we have got some good results. Though we doesn't get $r(3,10,40)$, but we do get some graphs only containing two 3-complete graph and no 10-independent sets, and we hope to find $r(3,10,40)$-graphs from these small fitness function graphs. And we can get some $r(3,10,39)$ graphs from these small fitness function graphs. After we analyze some of the graphs whose fitness function is 2 or 3, and we find there are 4 kinds of structure of 2 3-complete graph or 3 3-complete graph. There are four graphs, the first two containing 3 3-complete graphs, and the next two containing 2 3-compete graphs. Their adjacent matrix and the triangles are shown in Figure 2, of which the first two picture are the triangles contained in the first two matrixes.

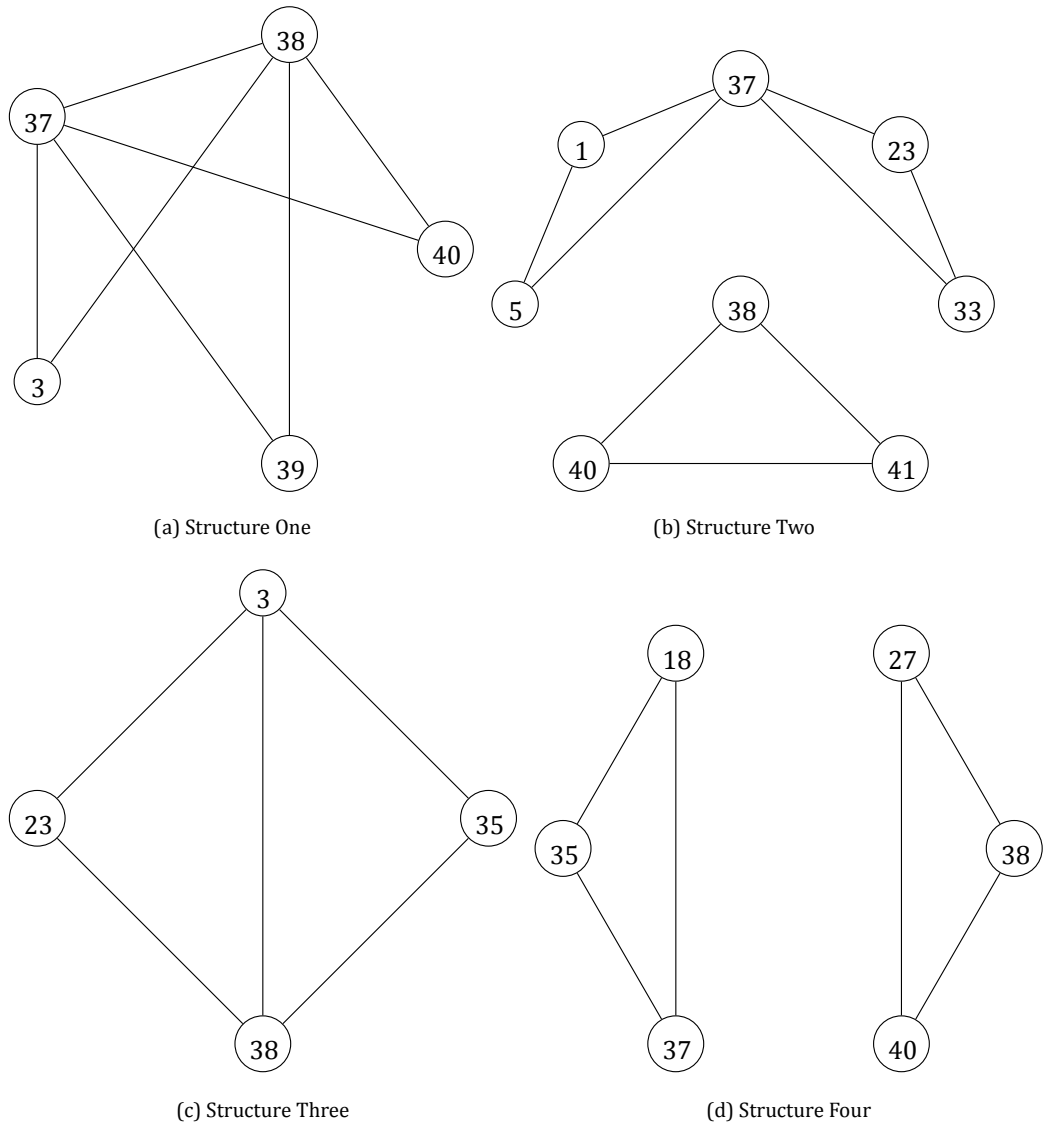

(a) Structure One  (b) Structure Two

(c) Structure Three  (d) Structure Four

Figure 5: four kinds of structures from the Fitness Graph

And the four kinds of graphs whose fitness function value is the best can be shown as following:

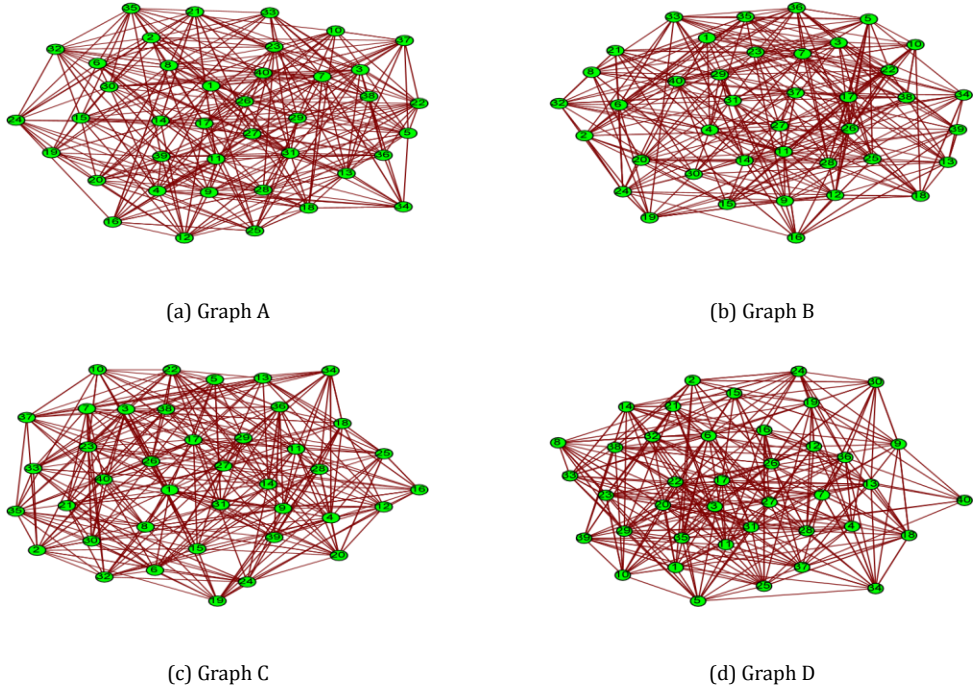

(a) Graph A  (b) Graph B

(c) Graph C  (d) Graph D

Figure 6: four *r*(3,9,39) Graphs Achieved by ABC Optimization

## 5. The Conclusion

In this article, visa Artificial Bee Colony to calculate the Ramsey Number R(3,10), we achieved four *r*(3,9,39) graphs at the first time to approximate inf R(3,10).

We also try to adjust some of the edges in one of these four graphs and always get another one of these four graphs or some graphs containing many 10-independent sets. So we strongly suspect that these four graphs maybe the best and *r*(3,10) = 40. We can see from the graphs that if we delete vertex 3 or 38 in graph-3 or delete vertex 37 or vertex 38 in graph-1 then we can get four *r*(3,9,39) graphs. They may be isomorphic to the one already found and maybe not. From these for the first time, four graphs we also try to add one more vertex into the graph and hope to construct a graph of *r*(3,10,40). The result shows these graphs cannot be extended into graph of r(3,10,40). So it is very likely that the present lower bound is the best lower bound for R(3,10).

In the near future, we will analyze whether four *r*(3,9,39) graphs are isomorphic or not. What's more, we may apply other artificial intelligence algorithm to the calculation of Ramsey number.

**6. Appendix**

As is pointed out in 4, we can get four *r*(3,9,40) graphs, and their matrix is shown below.

Graph A:
 1:2 3 4 5 6 7 8 9 36
2:1 10 20 25 27 31 32 34 39
3:1 15 23 24 26 30 32 35 37 38
4:1 13 21 26 29 30 33 35 40
5:1 11 18 24 27 30 32 34 38
6:1 12 22 25 28 31 33 34
7:1 14 19 28 29 31 33 35
8:1 17 18 21 22 30 33 34 37
9:1 16 19 20 23 31 32 35 37
10:2 11 12 16 18 19 24 30 38
11:5 10 15 17 22 23 26 35 37
12:6 10 14 17 20 21 27 32
13:4 14 15 17 18 19 28 31 39
14:7 12 13 16 22 23 25 34 37
15:3 11 13 16 20 21 29 33 36
16:9 10 14 15 17 26 27 28 40
17:8 11 12 13 16 24 25 29

18:5 8 10 13 25 26 29 35 40
19:7 9 10 13 25 26 27 34 40
20:2 9 12 15 24 26 28 30

21:4 8 12 15 24 25 28 31 39  
22:6 8 11 14 24 27 29 32 38  
23:3 9 11 14 27 28 29 33 36  
24:3 5 10 17 20 21 22 33 36  
25:2 6 14 17 18 19 21 30  
26:3 4 11 16 18 19 20 31 39  
27:2 5 12 16 19 22 23 35 37  
28:6 7 13 16 20 21 23 32 37  
29:4 7 15 17 18 22 23 34 39  
30:3 4 5 8 10 20 25 31 36  
31:2 6 7 9 13 21 26 30 40  
32:2 3 5 9 12 22 28 33 36  
33:4 6 7 8 15 23 24 32 39  
34:2 5 6 8 14 19 29 35  
35:3 4 7 9 11 18 27 34 36  
36:1 15 23 24 30 32 35 39 40  
37:3 8 9 11 14 27 28 38 39 40  
38:3 5 10 22 37 39 40  
39:2 13 21 26 29 33 36 37 38  
40:4 16 18 19 31 36 37 38  
 Graph B:  
1:2 3 4 5 6 7 8 9 37 38  
2:1 10 20 25 27 31 32 34 36  
3:1 15 23 24 26 30 32 35 40  
4:1 13 21 26 29 30 33 35 36  
5:1 11 18 24 27 30 32 34 36 37  
6:1 12 22 25 28 31 33 34 36  
7:1 14 19 28 29 31 33 35 36  
8:1 17 18 21 22 30 33 34 36  
9:1 16 19 20 23 31 32 35 36  
 10:2 11 12 16 18 19 24 30 39  
11:5 10 15 17 22 23 26 35 38 40

12:6 10 14 17 20 21 27 32 37
13:4 14 15 17 18 19 28 31 40
14:7 12 13 16 22 23 25 34 39
15:3 11 13 16 20 21 29 33 39
16:9 10 14 15 17 26 27 28 40
17:8 11 12 13 16 24 25 29 39
18:5 8 10 13 25 26 29 35 38
19:7 9 10 13 25 26 27 34
20:2 9 12 15 24 26 28 30
21:4 8 12 15 24 25 28 31
22:6 8 11 14 24 27 29 32 37
23:3 9 11 14 27 28 29 33 37
24:3 5 10 17 20 21 22 33 38
25:2 6 14 17 18 19 21 30 40
26:3 4 11 16 18 19 20 31 37
27:2 5 12 16 19 22 23 35 38
28:6 7 13 16 20 21 23 32 39
29:4 7 15 17 18 22 23 34
30:3 4 5 8 10 20 25 31
31:2 6 7 9 13 21 26 30 38
32:2 3 5 9 12 22 28 33 38
33:4 6 7 8 15 23 24 32 37
34:2 5 6 8 14 19 29 35 38
35:3 4 7 9 11 18 27 34 37
36:2 4 5 6 7 8 9 39 40
37:1 5 12 22 23 26 33 35 39 40
38:1 11 18 24 27 31 32 34 39 40
39:10 14 15 17 28 36 37 38
40:3 11 13 16 25 36 37 38

Graph C:
1:2 3 4 5 6 7 8 9 36
2:1 10 20 25 27 31 32 34 39

3:1 15 23 24 26 30 32 35 37 38
4:1 13 21 26 29 30 33 35 40
5:1 11 18 24 27 30 32 34 38
6:1 12 22 25 28 31 33 34
7:1 14 19 28 29 31 33 35
8:1 17 18 21 22 30 33 34 37
9:1 16 19 20 23 31 32 35 37
10:2 11 12 16 18 19 24 30 38
11:5 10 15 17 22 23 26 35 37
12:6 10 14 17 20 21 27 32
13:4 14 15 17 18 19 28 31 39
14:7 12 13 16 22 23 25 34 37
15:3 11 13 16 20 21 29 33 36
16:9 10 14 15 17 26 27 28 40
17:8 11 12 13 16 24 25 29
18:5 8 10 13 25 26 29 35 40
19:7 9 10 13 25 26 27 34 40
20:2 9 12 15 24 26 28 30
21:4 8 12 15 24 25 28 31 39
22:6 8 11 14 24 27 29 32 38
23:3 9 11 14 27 28 29 33 36 38
24:3 5 10 17 20 21 22 33 36
25:2 6 14 17 18 19 21 30
26:3 4 11 16 18 19 20 31 39
27:2 5 12 16 19 22 23 35 37
28:6 7 13 16 20 21 23 32 37
29:4 7 15 17 18 22 23 34 39
30:3 4 5 8 10 20 25 31 36
31:2 6 7 9 13 21 26 30 40
32:2 3 5 9 12 22 28 33 36
33:4 6 7 8 15 23 24 32 39
34:2 5 6 8 14 19 29 35

35:3 4 7 9 11 18 27 34 36 38  
36:1 15 23 24 30 32 35 39 40  
37:3 8 9 11 14 27 28 39 40  
38:3 5 10 22 23 35 39 40  
39:2 13 21 26 29 33 36 37 38  
40:4 16 18 19 31 36 37 38  

Graph D:  
1:2 3 4 5 6 7 8 9 40  
2:1 10 20 25 27 31 32 34 39  
3:1 15 23 24 26 30 32 35 38  
4:1 13 21 26 29 30 33 35 38  
5:1 11 18 24 27 30 32 34 36  
6:1 12 22 25 28 31 33 34 37  
7:1 14 19 28 29 31 33 35 38  
8:1 17 18 21 22 30 33 34 36  
9:1 16 19 20 23 31 32 35 38  
10:2 11 12 16 18 19 24 30 40  
11:5 10 15 17 22 23 26 35 38  
12:6 10 14 17 20 21 27 32 39  
13:4 14 15 17 18 19 28 31  
14:7 12 13 16 22 23 25 34 40  
15:3 11 13 16 20 21 29 33 39  
16:9 10 14 15 17 26 27 28  
17:8 11 12 13 16 24 25 29 37  
18:5 8 10 13 25 26 29 35 37 38  
19:7 9 10 13 25 26 27 34 37  
20:2 9 12 15 24 26 28 30 40  
21:4 8 12 15 24 25 28 31 37  
22:6 8 11 14 24 27 29 32 39  
23:3 9 11 14 27 28 29 33  
24:3 5 10 17 20 21 22 33  

25:2 6 14 17 18 19 21 30 36
26:3 4 11 16 18 19 20 31 39
27:2 5 12 16 19 22 23 35 38 40
28:6 7 13 16 20 21 23 32 36
29:4 7 15 17 18 22 23 34 36
30:3 4 5 8 10 20 25 31 37
31:2 6 7 9 13 21 26 30
32:2 3 5 9 12 22 28 33 40
33:4 6 7 8 15 23 24 32
34:2 5 6 8 14 19 29 35 38
35:3 4 7 9 11 18 27 34 36 37
36:5 8 25 28 29 35 39 40
37:6 17 18 19 21 30 35 39 40
38:3 4 7 9 11 18 27 34 39 40
39:2 12 15 22 26 36 37 38
40:1 10 14 20 27 32 36 37 38

## References

[1] E. W. Weisstein, Ramsey number.from mathworld–a wolfram web resource, [EB/OL], http://mathworld.wolfram.com/RamseyNumber.html.

[2] J. Goedgebeur, S. P. Radziszowski, New computational upper bounds for ramsey numbers r (3, k), Electronic Journal of Combinatorics 20 (2013) 301–311.

[3] P. Erdös, Some remarks on the theory of graphs, Bulletin of the American Mathematical Society 53 (4) (1947) 292–294.

[4] D. Conlon, A new upper bound for diagonal ramsey numbers, Annals of Mathematics (2009) 941–960.

[5] L. Haipeng, S. Wenlong, L. Zhenchong, The properties of selfcomplementary graphs and new lower bounds for diagonal ramsey numbers, Australasian Journal of Combinatorics 25 (2002) 103–116.

[6] J. H. Kim, The ramsey number r (3, t) has order of magnitude t2/log t, Random Structures & Algorithms 7 (3) (1995) 173–207.

[7] G. Exoo, M. Tatarevic, New lower bounds for 28 classical ramsey numbers, The Electronic Journal of Combinatorics 22 (3) (2015) P3–11.

[8] H. Harborth, S. Krause, Ramsey numbers for circulant colorings, Congressus Numerantium (2003) 139–150.

[9] H. Fujita, A new lower bound for the ramsey number r (4, 8), arXiv preprint arXiv:1212.1328.

[10] S. Radziszowski, D. Kreher, Upper bounds for some ramsey numbers r (3, k) , The Journal of Combinatorial Mathematics and Combinatorial Computing (1988) 207–212.

[11] G. Exoo, Announcement-on the ramsey numbers r (4, 6), r (5, 6) and r (3, 12), Ars Combinatoria 35 (1993) 85–85.

[12] G. Exoo, On two classical ramsey numbers of the form r(3,n), SIAM Journal [415] on Discrete Mathematics 2 (4) (1989) 488–490.

[13] G. Exoo, A lower bound for schur numbers and multicolor ramsey numbers of k3, Electron. J. Combin 1 (1994) R8.

[14] J. R. Griggs, An upper bound on the ramsey numbers r (3, k), Journal of Combinatorial Theory, Series A 35 (2) (1983) 145–153.

[15] H.-p. Luo, W.-l. Su, Y.-Q. Shen, New lower bounds of ten classical ramsey numbers, Australasian Journal of Combinatorics 24 (2001) 81–90.

[16] A. Robertson, New lower bounds for some multicolored ramsey numbers, Journal of Combinatorics 6 (1998) 51–56.

[17] D. Samana, V. Longani, Upper bounds of ramsey numbers, Applied Math[425]ematical Sciences 6 (98) (2012) 4857–4861.

[18] H. Haanp¨a¨a, A lower bound for a ramsey number, Congr. Numer 144 (2000) 189–191.

[19] R. E. Greenwood, A. M. Gleason, Combinatorial relations and chromatic graphs, Canad. J. Math 7 (1) (1955) 7.

[20] J. E. Graver, J. Yackel, Some graph theoretic results associated with ramsey's theorem, Journal of Combinatorial Theory 4 (2) (1968) 125–175.

[21] J. G. Kalbfleisch, Chromatic graphs and Ramsey's theorem, University of Waterloo, 1966.

[22] B. D. McKay, Z. K. Min, The value of the ramsey number r (3, 8), Journal [435] of Graph Theory 16 (1) (1992) 99–105.

[23] C. M. Grinstead, S. M. Roberts, On the ramsey numbers r (3, 8) and r (3, 9), Journal of Combinatorial Theory, Series B 33 (1) (1982) 27–51.

[24] G. Exoo, A lower bound for r (5, 5), Journal of graph theory 13 (1) (1989) 97–98.

[25] Exoo.Geoffrey, On the ramsey number r(4, 6), Electronic Journal of Combinatorics 19 (1) (2012) 1435–1447.

[26] B. D. Mckay, K. Piwakowski, S. P. Radziszowski, Ramsey numbers for triangles versus almost-complete graphs., Ars Combinatoria -Waterloo then Winnipeg- 73.

[27] B. D. McKay, S. P. Radziszowski, Subgraph counting identities and ramsey numbers, journal of combinatorial theory, Series B 69 (2) (1997) 193–209.

[28] P. Dirac, The lorentz transformation and absolute time, Physica 19 (1-12) (1953) 888–896. doi:10.1016/S0031-8914(53)80099-6.

[29] R. Feynman, F. Vernon Jr., The theory of a general quantum system interacting with a linear dissipative system, Annals of Physics 24 (1963)

[30] S. P. Radziszowski, D. L. Kreher, Search algorithm for ramsey graphs by union of group orbits, Journal of Graph Theory 12 (1) (1988) 59–72.

[31] G. Exoo, Some new ramsey colorings, JOURNAL OF COMBINATORICS [455] 5 (1998) 427–432.

[32] A. Lesser, Theoretical and computational aspects of ramsey theory, Examensarbeten i Matematik, Matematiska Institutionen, Stockholms Universitet 3.

[33] G. Exoo, Appling optimization algorithm to ramsey problems, Graph Theory, Combinatorics, Algorithms and Applications (Philadelphia)(Y. Alavi, ed.), SIAM (1989) 175–179.

[34] K. Piwakowski, Applying tabu search to determine new ramsey numbers, Electronic J. Combinatorics 3 (1996) 1–4.

[35] S. P. Radziszowski, et al., Small ramsey numbers, Electron. J. Combin DS1 (2014).

[36] X. Xiaodong, X. Zheng, G. Exoo, S. P. Radziszowski, et al., Constructive lower bounds on classical multicolor ramsey numbers, JOURNAL OF COMBINATORICS 11 (2) (2004) R35.

[37] M. K. Apalak, D. Karaboga, B. Akay, The artificial bee colony algorithm inlayer optimization for the maximum fundamental frequency of symmetrical laminated composite plates, Engineering Optimization 46 (3) (2014) 420–437.

[38] D. Karaboga, B. Gorkemli, C. Ozturk, N. Karaboga, A comprehensive survey: artificial bee colony (abc) algorithm and applications, Artificial Intelligence Review 42 (1) (2014) 21–57.

[39] I. Develi, Y. Kabalci, A. Basturk, Artificial bee colony optimization for modelling of indoor plc channels: A case study $_{480}$ from turkey, Electric Power Systems Research 127 (2015) 73–79.

[40] D. Karaboga, B. Basturk, Artificial bee colony (abc) optimization algorithm for solving constrained optimization problems, in: Foundations of Fuzzy Logic and Soft Computing, Springer, 2007, pp. 789–798.

[41] D. Karaboga, C. Ozturk, A novel clustering approach: Artificial bee colony (abc) algorithm, Applied soft computing 11 (1) (2011) 652–657.

[42] F. Gao, F. xia Fei, Y. fang Deng, Y. bo Qi, B. Ilangko, A novel non-lyapunov approach through artificial bee colony algorithm for

detecting unstable periodic orbits with high orders, Expert Systems with Applications 39 (16) (2012) 12389 – 12397.

[43] F. Gao, Y. bo Qi, , Q. Yin, J. Xiao, An novel optimal pid tuning and on–line tuning based on artificial bee colony algorithm, in: The 2010 International Conference on Computational Intelligence and Software Engineering (CiSE2010), IEEE, Wuhan, China, 2010, pp. 425–428.

[44] F. Gao, Y. bo Qi, , Q. Yin, J. Xiao, Online synchronization of uncertain chaotic systems by artificial bee colony algorithm in a non–lyapunov way, in: The 2010 International Conference on Computational Intelligence and Software Engineering (CiSE 2010), IEEE, Wuhan, China, 2010, pp. 1–4.

[45] F. Gao, Y. bo Qi, , Q. Yin, J. Xiao, An artificial bee colony algorithm for unknown parameters and time–delays identification of chaotic systems, in: the Fifth International Conference on Computer Sciences and Convergence Information Technology (ICCIT10), IEEE, Seoul, Korea, 2010, pp. 659 – 664.

[46] F. Gao, Y. bo Qi, , Q. Yin, J. Xiao, A novel non–lyapunov approach in discrete chaos system with rational fraction control by artificial bee colony algorithm, in: 2010 International Conference on Progress in Informatics and Computing(PIC-2010), IEEE, Shanghai, China, 2010, pp. 317 – 320.

[47] G. Zhu, S. Kwong, Gbest-guided artificial bee colony algorithm for numerical function optimization, Applied Mathematics and Computation 217 (7) (2010) 3166–3173.

[48] D. Karaboga, B. Akay, C. Ozturk, Artificial bee colony (abc) optimization algorithm for training feed-forward neural networks, in: Modeling decisions for artificial intelligence, Springer, 2007, pp. 318–329.

[49] B. Akay, D. Karaboga, A modified artificial bee colony algorithm for real[520] parameter optimization, Information Sciences 192 (2012) 120–142.

[50] D. Karaboga, B. Basturk, A powerful and efficient algorithm for numerical function optimization: artificial bee colony (abc) algorithm, Journal of global optimization 39 (3) (2007) 459–471.

[51] D. Karaboga, B. Basturk, On the performance of artificial bee colony (abc) [525] algorithm, Applied soft computing 8 (1) (2008) 687–697.

[52] D. Karaboga, B. Akay, A comparative study of artificial bee colony algorithm, Applied Mathematics and Computation 214 (1) (2009) 108–132.